\pdfoutput=1

\documentclass[sn-apa]{sn-jnl}

\usepackage[FIGTOPCAP]{subfigure}
\usepackage{multirow}
\usepackage{lscape}
\usepackage{tabularx}
\usepackage{rotating}
\usepackage{tabu}
\usepackage{longtable}[=v4.13]
\usepackage{booktabs}
\usepackage{apacite}
\usepackage{amsmath}
\usepackage{enumitem}

\jyear{2021}%

\begin{document}

\title[Online AutoML: An adaptive AutoML framework for online learning]{Online AutoML: An adaptive AutoML framework for online learning}


\author*[1]{\fnm{Bilge} \sur{Celik}}\email{B.Celik.Aydin@tue.nl} 

\author*[1]{\fnm{Prabhant} \sur{Singh}}\email{P.Singh@tue.nl}

\author*[1]{\fnm{Joaquin} \sur{Vanschoren}}\email{J.Vanschoren@tue.nl}

\affil[1]{\orgdiv{Department of Computer Science \& Mathematics}, \orgname{Eindhoven University of Technology}, \orgaddress{\street{Groene Loper 5}, \city{Eindhoven}, \postcode{5600MB}, \country{The Netherlands}}}


\abstract{}
Automated Machine Learning (AutoML) has been used successfully in settings where the learning task is assumed to be static. In many real-world scenarios, however, the data distribution will evolve over time, and it is yet to be shown whether AutoML techniques can effectively design online pipelines in dynamic environments. This study aims to automate pipeline design for online learning while continuously adapting to data drift. For this purpose, we design an adaptive Online Automated Machine Learning (OAML) system, searching the complete pipeline configuration space of online learners, including preprocessing algorithms and ensembling techniques. This system combines the inherent adaptation capabilities of online learners with fast automated pipeline (re)optimization. Focusing on optimization techniques that can adapt to evolving objectives, we evaluate asynchronous genetic programming and asynchronous successive halving to optimize these pipelines continually. We experiment on real and artificial data streams with varying types of concept drift to test the performance and adaptation capabilities of the proposed system. The results confirm the utility of OAML over popular online learning algorithms and underscore the benefits of continuous pipeline redesign in the presence of data drift.




\maketitle

\section{Introduction}\label{Intro}
Machine learning has moved beyond static environments with the rise of streaming data sources in every corner of life. This new era of data also introduced new challenges that created a need to redefine old solutions. Online or stream learning addresses the research questions arising from this transformation \citep{Gomes2019}. A more dynamic environment naturally entails change, time constraints, and uncertainty. One of the main challenges of online learning is successful and timely adaptation to a change in data-generating dynamics, known as concept drift \citep{Gama2014}. Machine learning algorithms are usually also bounded by limited memory and can only do a single pass over the data samples in pursuit of this goal. Hence, online algorithms have emerged that cope with all these challenges. However, these algorithms also come with hyperparameters that need to be tuned to the task at hand and, since these tasks are dynamic in nature, they may need to be retuned when concept drift occurs. Recent studies focus on online hyperparameter tuning and algorithm selection \citep{veloso2018, Carnein2019}. This also extends to the data preprocessing techniques that need to be selected, tuned, and continuously adapted. The rising interest in automated algorithm adaptation to underlying changes in the data highlights the importance of updating both the model parameters and hyperparameters and the need for a flexible strategy to do so \citep{bakirov2021}. 

Automated machine learning (AutoML) has demonstrated its benefits in hyperparameter tuning and algorithm selection in several machine learning scenarios \citep{Thornton2013, Feurer2015}, maturing into a field with many alterations for batch data and extensive comparative studies  \citep{Gijsbers2019}. Some of the initial attempts in carrying these capabilities into the online world make use of well-known AutoML libraries, adapted to data stream settings by re-optimizing pipelines as needed \citep{Madrid2019, Celik2021}. Separately, initial steps are taken to design AutoML systems that automatically configure online learning algorithms \citep{Wu2021}. However, the latter still lack fundamental AutoML features such as exploring a large space of learning algorithms and including preprocessing steps in pipelines.

In this paper, we aim to combine the power of AutoML approaches to operate over a wide gamut of pipeline configurations with the intrinsic adaptability of online learning algorithms. We design a search space centered on online learning algorithms, ensembles, and preprocessors. The search spans pipelines with one or more steps, focusing on algorithm selection and hyperparameter optimization simultaneously. Available optimization algorithms include but are not limited to asynchronous evolutionary algorithm and asynchronous successive halving, due to their ability to adapt to evolving objectives \citep{Celik2021}. Our methods are integrated into a modular AutoML system \citep{Gijsbers2021}, thus allowing further extensions of the search space, optimizers, and objective functions. Moreover, to ensure fast adaptation to concept drift, our system introduces novel components such as backup ensembles and model stores. This Online AutoML framework (OAML) is, to the best of our knowledge, the first to propose a flexible and practical AutoML system for adaptive online learning pipelines. 

We evaluate our system on a range of concept drift data with different concept drift characteristics, and compare against popular adaptive learners. Our findings indicate that optimizing complete pipelines works best for fast adaptation to concept drift and that the system effectively leverages intermediate pipeline evaluations and updates.

The remainder of the paper is organized as follows. Section~\ref{Problem} formulates our core problem and its intrinsic challenges. Following that, Section~\ref{background} provides the necessary background on AutoML and Online Adaptive Learning. Section~\ref{Existing} introduces existing approaches with a similar goal to ours. Our proposed method, OAML, is detailed in Section~\ref{OAML}. Section~\ref{experimentdesign} describes the design of our empirical evaluation, and we present and analyze the results in Section~\ref{sub:results}. Section~\ref{Concl} concludes.

\section{Problem definition}\label{Problem}
Finding the optimal configuration of machine learning pipelines is one of the main goals of AutoML research. In the context of batch learning, the problem is described for a fixed dataset $\boldsymbol{D} = \big\{(\boldsymbol{x}^i, y^i)$, $i=1,..,n\big\}$. \textit{Combined algorithm selection and hyperparameter optimization (CASH)} \citep{Thornton2013} is the search over learning algorithms $\boldsymbol{A}$ and associated hyperparameter spaces $\boldsymbol{\Lambda_A}$ for an optimal combination $A^*_{\lambda^*}$ that maximizes the performance of prediction over $k$ subsets of $\boldsymbol{D}$ (e.g., $k$ cross-validation folds, yet it can also be formulated with holdout evaluation). Equation \ref{eq:1} formalizes this optimization problem, where $L$ is an evaluation measure, and $\big\{\boldsymbol{X}_{tr}, \boldsymbol{y}_{tr}\big\}$ and $\big\{\boldsymbol{X}_{val}, \boldsymbol{y}_{val}\big\}$ represent the training and validation sets, respectively. The search can be extended to include preprocessing algorithms as well as postprocessing steps, in which case $\boldsymbol{A}$ is the space of all possible pipelines applicable to the specific machine learning problem.

\begin{equation}
\begin{split}
A^*_{\lambda^*} =
\operatorname*{argmin}_{%
       \substack{%
         \forall A^j \in \boldsymbol{A} \\
         \forall \lambda \in \boldsymbol{\Lambda_A}
       }
     }
\frac{1}{k} 
\sum_{f=1}^{k} L \left( A^j_\lambda, \big\{\boldsymbol{X}^f_{tr}, \boldsymbol{y}^f_{tr}\big\}, \big\{\boldsymbol{X}^f_{val}, \boldsymbol{y}^f_{val}\big\} \right)
\end{split}
\label{eq:1}
\end{equation}

However, in this study, the exceptions  compared to the original problem are the temporal dimension of data and that it is considered to be infinitely long. Some other constraints online learning imposes are the requirement to process the data in order of arrival and restricting the memory usage to a limited scale \citep{Gama2014}. AutoML for online learning can be described similarly to the batch learning setting when the online system allows for multi pass of data and a certain amount of memory to allow for the adaptation of AutoML search algorithms. Single pass with zero memory systems would require further changes to the problem description of CASH. Hence, not all data can be stored in memory, which limits the range of possible algorithms to a set $\boldsymbol{A}_{OL} \subset \boldsymbol{A}$. In addition, evaluation happens in a prequential way, where $\boldsymbol{X}_{t}$ is the batch (or window) of data at time step $t$. Here we count time steps between the batches of data, yet data can still arrive in single samples. As shown in Equation \ref{eq:2}, our objective is now to return the best \textit{online} pipeline $A^*_{\lambda,t}$ at each time step $t$, where this pipeline is continually trained on the previous batch $\big\{\boldsymbol{X}_{t-1}, \boldsymbol{y}_{t-1}\big\}$ and evaluated on the current one $\big\{\boldsymbol{X}_{t}, \boldsymbol{y}_{t}\big\}$:

\begin{equation}
\begin{split}
A^*_{\lambda, t} =
\operatorname*{argmin}_{%
       \substack{%
         \forall A^j \in \boldsymbol{A}_{OL} \\
         \forall \lambda \in \boldsymbol{\Lambda}
       }
     }
L \left( A^j_{\lambda, t}, \big\{\boldsymbol{X}_{t-1}, \boldsymbol{y}_{t-1}\big\}, \big\{\boldsymbol{X}_{t}, \boldsymbol{y}_{t}\big\} \right)
\end{split}
\label{eq:2}
\end{equation}

Another big challenge of online learning are unpredictable shifts in the data distribution resulting from data generating processes, known as \textit{concept drift}. Although this shift can happen in several components of the underlying data distribution, the most interesting and challenging case for supervised learning is a change in the posterior probabilities of output variables, $p_{t}(y \mid X)$ at a certain time step t. This is known as \textit{real concept drift} (Equation \ref{eq:3}) and affects the class boundary, thus requiring the learner to adapt to the change. Since real concept drift can occur without affecting the input data distribution, $p(X)$, it is mostly detected through changes in the predictive performance of the learner. 

\begin{equation}
\exists X: p_{t_0} (y \mid X) \neq p_{t_1} (y \mid X)
\label{eq:3}
\end{equation}

Concept drift can occur in many forms with different characteristics, where the duration, transition and magnitude of the change are expected to have the biggest effect on a learner's ability to adapt \citep{Webb2016}. Therefore, in this research we will evaluate our methods on data with both \textit{abrupt} and \textit{gradual drift}, and typically with high drift magnitudes, which are most challenging. Abrupt drift occurs when the concept changes suddenly and the duration of this shift is smaller than a certain time period, usually over a single sample. Gradual drift, on the other hand, occurs when the difference between concepts (i.e., the drift magnitude) over a time period is smaller than a maximum value. 

\section{Background}\label{background}
\subsection{AutoML}\label{Automl}
Different approaches to the CASH problem in combination with a variety of 
pipeline structures and search spaces have a vast amount of AutoML systems. Most differ mainly in the optimization algorithm used to search the pipeline configuration space. \textit{Bayesian optimization (BO)} \citep{Thornton2013, Feurer2015}, one of the most widely and successfully used methods in offline settings, fits a probabilistic surrogate model over the evaluated pipelines in the search space and predicts the performance and uncertainty of unseen configurations. Gaussian Processes are one of the most popular choices for the surrogate model for smaller hyperparameter search spaces \citep{Snoek2012} while Random Forests are often used for larger spaces \citep{Feurer2015}. \textit{Evolutionary methods} are another effective approach for pipeline optimization \citep{Olson2016b, Gijsbers2021}, in which pipelines are evolved with crossover and mutations through genetic programming. GAMA (General Automated Machine Learning Assistant) \citep{Gijsbers2021} is an AutoML library that uses asynchronous genetic programming. This approach has also shown to be effective for adaptive learning with online data \citep{Celik2021}. Hence, we use GAMA's genetic programming configuration and search algorithms as one of our optimization methods, as explained in more detail in Section \ref{OAML}. The library also includes other search algorithms such as \textit{Random Search}, which randomly samples hyperparameter configurations, and \textit{Asynchronous Successive Halving (ASHA)}, which speeds up random search by asynchronous early stopping.

\subsection{Online adaptive learning}\label{OnlineLearning}
What distinguishes online learning from offline learning is the progress of data availability. In online learning, data is received over time and used in arriving order for updating the model \citep{Gama2014}. This is one of the main challenges for AutoML in online learning since existing search algorithms require access to all data a priori.
The online model can have partial access to previous data in case the model keeps a limited memory \citep{Maloof2004}. In general, though, the assumption is that past and future training data is unavailable to the learner and each data sample only has a single pass through the training of the learner. Another characteristic of online learning is anytime prediction: the trained learner can be used for prediction at any given time. 

Adaptation, updating the learning model or pipeline by retraining and/or re-tuning, is required when data evolves over time and results in concept drift, making the previous model obsolete. In \textit{online adaptive learning}, the continuous cycle includes prediction, evaluation and training steps. Whether adaptation is required or not is determined in the evaluation step, where a drift detector checks whether the current model suffers a negative performance change. Some well-known drift detectors keep track of variables related to model performance, while others use a data window approach. DDM (Drift Detection Method) records the error-rate of the learner and fits a distribution over time \citep{Gama2004}. It emits a drift alert when the confidence interval exceeds a certain threshold. Another benchmark method, EDDM (Early Drift Detection Method), monitors the distance between the errors in addition to their frequency \citep{BaenaGarcia2006}. Both methods work well with abrupt concept drift and have low memory footprints, yet EDDM is known to be superior in detecting gradual drift. One drawback of EDDM is the occurence of false alarms in the early stages of learning due to the small distance between initial errors. ADWIN (Adaptive Sliding Window), a well-known window approach, compares the means of two subsets of data and emits a drift signal when there is a significant difference between those means \citep{Bifet2007}. ADWIN requires more memory and execution runtime compared to error-rate monitoring methods, due to the need to maintain sliding windows. 

When drift is detected at a certain time point, these methods can adapt the model locally or globally depending on their adaptation strategy and the characteristics of the drift \citep{Gama2014}. Discriminant classifiers or naive Bayes methods require model replacement, i.e., delete the old model and train a completely new one. Yet, in some occasions the drift characteristics can allow a model adaptation to restore the performance of the learner. Decision trees allow that kind of adaptation due to their modular structure. \cite{Celik2021} show that drift characteristics such as the magnitude and duration of the drift influence the correct adaptation strategy to follow. In case the learner keeps a limited memory of past samples, a forgetting mechanism is another critical aspect to adapt successfully in concept drift scenarios. A straightforward and common approach is a constant rate sliding window where data samples are erased and added at the same rate as data flows. Smaller sliding windows contribute to faster adaptation to the new concept, yet can lead to unstable and low performance due to insufficient training data. Dynamic weighing of data is another possible forgetting mechanism, where the dynamic rate can be adjusted in case of drift. 

\textit{Prequential evaluation}, also known as interleaved test-then-train, is the most widely used evaluation approach, where each individual sample is first used to evaluate the performance and then to update the learner. This approach can also be applied in batches of arriving samples (\textit{data chunk evaluation}). Prequential accuracy is dynamic as the performance is updated incrementally. 
This contributes to the difficulty of using AutoML for online learning, since most AutoML libraries use cross-validation, which can't be applied here. Holdouts can be used if the temporal order of the data is respected.
Although prequential evaluation scores are shown to be more pessimistic compared to holdout evaluation \citep{Gama2013}, it is widely used in adaptive online learning methods. 

Among online learning methods, online bagging approaches are among the best performing ones due to the advantages ensembles bring to smooth drift adaptation. Oza Bagging \citep{Oza2001} and its updated version Leveraging Bagging \citep{Bifet2010} 
simulate re-sampling in online settings. In the former, ensemble diversity is obtained by increased randomization of the data, yet it also brings a heavier computational burden. Another popular ensemble approach following a re-sampling strategy is the Adaptive Random Forest (ARF) \citep{Gomes2017}. ARF also includes random feature re-sampling for splitting the nodes, which contributes to diversity. Each base tree is monitored and retrained individually in case of drifts. In order to reduce response times in the adaptation process, base trees are trained in the background when the detector gives a warning of a possible drift. These adaptive features make ARF one of the most well-performing online learning methods. Among the non-ensemble learners, the Hoeffding Adaptive Tree \citep{Domingos2000} is one of the fastest adapting approaches. It uses a drift detector to monitor and update individual branches.

\section{Related Work}\label{Existing}
Automating pipeline configuration in data streams gained interest over the last few years with the increase of online learning use cases. Some research focuses on hyperparameter tuning under concept drift, yet they restrict the optimization problem to a single learner. For stream clustering, \textit{confStream} \citep{Carnein2019} keeps an ensemble of several hyperparameter configurations of a stream learning algorithm, and uses their individual performances to train a linear regression model that predicts which new configurations to add. The evaluation shows an improvement over default clustering algorithm. \textit{SSPT (Single-pass Self Parameter Tuning)} \citep{veloso2018} is another auto-hyperparameter tuning method that uses a heuristic search algorithm. The approach is problem agnostic and again designed for a single learner and two hyperparameters.

Another line of research focuses on adapting a previously trained stream learning algorithm in case of concept drift, by automatically selecting an appropriate adaptation strategy. This approach is extended with automated adaptations of several ensemble stream learners \citep{bakirov2021}. However, these adaptation strategies can only be applied to a single model.

A step from stream learning adaptation towards AutoML is taken by extending existing AutoML libraries with several adaptation strategies, often based on drift detection, that allow them to retrain or re-optimize models in online learning scenarios \citep{Celik2021}. These adaptive AutoML methods perform better than stream learning algorithms across many tasks with various drift characteristics. Moreover, several AutoML techniques are used to examine the effectiveness of different search algorithms (e.g., Bayesian Optimization or Evolutionary techniques) in adapting to concept drift. Likewise, Madrid et al. \citep{Madrid2019} extend Autosklearn with a drift detector and two adaptation algorithms. The results corroborate the potential of AutoML for stream learning settings. However, in both studies, the underlying search spaces contain batch learning algorithms designed for offline settings, instead of the online learning algorithms considered in this work. 

\textit{ChaCha (Champion-Challengers)} \citep{Wu2021} is one of the most similar works to our purpose as it uses an AutoML setting designed for online learning. The search space of configurations is expanded progressively based on the online performance of existing base learners. It also balances computational effort by categorizing choices based on their learning cost and assigning resources only to the most promising ones. ChaCha is built on the FLAML \citep{Wu2021} library and uses algorithms from the Vowpal Wabbit online machine learning library. The method considers only one base learning algorithm at a time and focuses on finding promising hyperparameter settings for it. Therefore, it supports neither the optimization of complete pipelines (with preprocessing), nor the combined algorithm selection and hyperparameter optimization problem typical of AutoML.  To the best of our knowledge, our online OAML system is the first to propose an automated system for tuning and selecting online learning algorithms to create full pipelines including data and feature preprocessing. 

\section{Online AutoML (OAML)}\label{OAML}
In this paper, we introduce an automated adaptive online learning method, \textit{OAML (Online Automated Machine Learning)}, that is developed to solve the online CASH problem described in Section \ref{Problem}. Currently it only supports classification tasks, yet it can easily be extended to other supervised machine learning problems in the future. The model search space comprises a large set of online learning classifiers, ensembles and preprocessors, all implemented in the online learning library River \citep{River2020}, which is described further in Section~\ref{sub:searchspace}. 

Pipelines can include one or more of these algorithms, including data and feature preprocessing steps. It performs an online model search that combines algorithm selection and hyperparameter configuration, similar to offline versions of AutoML libraries. To the best of our knowledge, this is the only automated online learning system that includes a search space combining multiple online algorithms, preprocessors, as well as their hyperparameters. OAML can be constrained with a time budget for the pipeline design and the optimized pipelines are used in the online learning phase. A constant-rate sliding window approach is used to manage memory in a restricted way and also provide an up-to-date training set for pipeline search. OAML uses prequential evaluation to validate online pipelines according to a user-selected metric. It is publicly available, integrated into the open source AutoML library GAMA.\footnote{\url{https://github.com/openml-labs/gama/tree/oaml}}

\subsection{Search space design}\label{sub:searchspace}

The classifier algorithms in the search space mostly focus on adaptive ensemble methods since they are shown to be the most successful for online learning under concept drift \citep{Gama2014}. Our selection includes \textit{Oza Bagging} \citep{Oza2001} with the ADWIN drift detector, \textit{Leveraging Bagging} \citep{Bifet2010}, \textit{Ada Boosting} \citep{Oza2001}, and \textit{Adaptive Random Forests} \citep{Gomes2017}. The base learners to be considered in the ensemble methods are online versions of \textit{Logistic Regression}, \textit{k-Nearest Neighbors (KNN) Classifiers}, \textit{Perceptrons} and \textit{Hoeffding Trees}. The base learners are chosen without adaptive mechanisms since each ensembling method includes a drift detector. The \textit{Hoeffding Adaptive  Tree (HAT)} \citep{Hulten2001} is also added as an independent single learner due to its efficiency.

The search space also includes online versions of preprocessing algorithms, including data scaling and normalization, categorical variable encoding and feature extraction. The selected methods are \textit{Adaptive Standard Scaler}, \textit{Binarizer},  \textit{Maximum Absolute Scaler}, \textit{MinMax Scaler}, \textit{Normalizer}, \textit{Robust Scaler}, \textit{Standard Scaler} and \textit{Polynomial Feature Extender}. We also aim to include further preprocessing techniques for missing value imputation and feature selection.\footnote{Some preprocessors could not yet be included since their implementation in River still contained bugs. We are collaborating with the River developers to resolve these and extend the search space further.} 

The details of these algorithms can be found in \cite{River2020}. A list of the main hyperparameters for all these methods, with their defaults and value ranges, is shown in Table~\ref{tab:searchspace}. As mentioned earlier, there is limited research on hyperparameter optimization in this setting, and the existing work only scarcely explored hyperparameter grids. Hence, there is no go-to reference for these search intervals, and we had to design these based on our own insight and application defaults.

\begin{sidewaystable}[!]
\caption{Search space of OAML, with the online learners on top, and the main preprocessing methods below.}
\label{tab:searchspace}
\resizebox{0.9\linewidth}{!}{%
\begin{tabu}{llll} 
\toprule
\textbf{Model} & \textbf{Hyperparameter} & \textbf{Default value} & \textbf{Search range} \\ 
\hline
\multirow{8}{*}{Hoeffding Adaptive Tree (HAT)} & grace\_period & 200 & {[}50 - 350] \\
 & split\_criterion & info\_gini & \{gini, hellinger, info\_gini\} \\
 & split\_confidence & 1.00E-07 & \{1.00E-02, 1.00E-04, 1.00E-07, 1.00E-09\} \\
 & tie\_threshold & 0.05 & \{0.02, 0.03, 0.04, 0.05, 0.06, 0.07, 0.08\} \\
 & leaf\_prediction & nba & \{mc, nb, nba\} \\
 & bootstrap\_sampling & True & \{True, False\} \\
 & drift\_window\_threshold & 300 & \{100, 200, 300, 400, 500\} \\
 & adwin\_confidence & 2.00E-03 & \{2.00E-04, 2.00E-03, 2.00E-02\} \\ 
\hline
\multirow{8}{*}{Adaptive Random Forest (ARF)} & n\_models & 10 & {[}1-20] \\
 & max\_features & sqrt (n\_features) & \{0.2, 0.5, 0.7, 1.0, sqrt(n\_features),~ log2(n\_features), None\} \\
 & lambda\_value & 6 & {[}2 - 10] \\
 & grace\_period & 50 & {[}50 - 350] \\
 & split\_confidence & 1.00E-02 & \{1.00E-09, 1.00E-07, 1.00E-04, 1.00E-02\} \\
 & tie\_threshold & 0.05 & \{0.02, 0.03, 0.04, 0.05, 0.06, 0.07, 0.08\} \\
 & leaf\_prediction & nba & \{mc, nb, nba\} \\
 & nb\_threshold & 0 & \{0, 10, 20, 30, 40, 50\} \\ 
\hline
\multirow{5}{*}{Leveraging Bagging} & model & - & \{LogisticRegression, KNNClassifier, Perceptron, HoeffdingTreeClassifier\} \\
 & n\_models & 10 & {[}1 - 20] \\
 & w & 1 & {[}1 - 10] \\
 & adwin\_delta & 0.002 & \{0.001, 0.002, 0.005, 0.01\} \\
 & bagging\_method & bag & \{bag, me, half, wt, subag\} \\ 
\hline
\multirow{2}{*}{ADWIN (Oza) Bagging} & model & - & \{LogisticRegression, KNNClassifier, Perceptron, HoeffdingTreeClassifier\} \\
 & n\_models & 10 & {[}1 - 20] \\ 
\hline
\hline
\multirow{4}{*}{Robust Scaler} & with\_centering & True & \{True, False\} \\ 
\cline{4-4}
 & with\_scaling & True & \{True, False\} \\
 & q\_inf & 0.25 & {[}0 - 1, +0.05] \\
 & q\_sup & 0.75 & {[}0 - 1, +0.05] \\ 
\hline
Standard Scaler & - & - & - \\
Adaptive Standard Scaler & - & - & - \\
MaxAbs Scaler & - & - & - \\
MinMax Scaler & - & - & - \\
Normalizer & order & L2 norm & \{L1, L2\} \\
Binarizer & threshold & 0 & {[}0 - 1.01, +0.05] \\ 
\hline
\multirow{3}{*}{Polynomial Extender} & degree & 2 & \{2, 3\} \\
 & interaction\_only & False & \{True, False\} \\
 & include\_bias & False & \{True, False\} \\
\bottomrule
\end{tabu}
}
\end{sidewaystable}

\begin{figure}[t]
    \centering
    \includegraphics[scale=0.4]{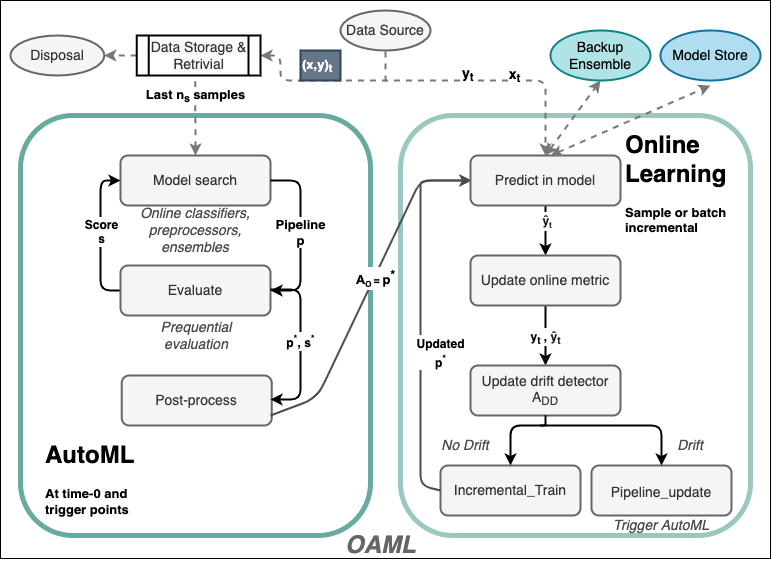}
    \caption{OAML Framework (Described in Section - \ref{OAML-general})}
    \label{fig:oaml}
\end{figure}

\subsection{Method overview}\label{OAML-general}
Figure~\ref{fig:oaml} shows an overview of the structure of our method, with the different system modules and flow. A more formal description in pseudo-code is given in Section \ref{OAML-ensemble}. The online learning process begins with an initial AutoML search (shown on the left of Fig.~\ref{fig:oaml}) using an initial batch of $n_0$ samples $(\boldsymbol{x_0}, ..., \boldsymbol{x_{n_0}})$ of a data stream. The search algorithm $S$ trains and evaluates pipeline configurations with a classification metric $M_a$ over the initial batch of data within the given time budget $t_{max}$. The best-found single pipeline of the search, $P^*_{0}$, is fitted to the available data and the trained pipeline is passed to the online learning module (shown on the right of Fig.~\ref{fig:oaml}) to be assigned as the online model $A_O$. From now on, data samples are assumed to arrive one by one, hence $P^*_{0}$ is used to predict the label for $\boldsymbol{x_t}$ as $\hat{y_t}$. When the real label $y_t$ is known to the model, the online evaluation metric $M_o$ is updated with the feedback.\footnote{The AutoML evaluation metric $M_a$ and online metric $M_o$ are usually the same, but could potentially be defined differently. For instance, $M_a$ could be a cheaper approximation to speed up the search.} At this point, $(y_t, \hat{y_t})$ is also used to update the drift detector, $A_{DD}$. In case of a drift signal, OAML is triggered to start a new AutoML pipeline search, possibly in parallel, with the batch of the latest $n_s$ samples, where $n_s$ is the sliding window size. This sliding window allows the search algorithm to have a restricted memory and discard outdated samples. This trigger, as shown in Figure~\ref{fig:oaml} at the \textit{Drift} branch, restarts AutoML \textit{Search} and updates the online pipeline, $A_O$. \textit{Pipeline update} phase in Online Learning corresponds to the new AutoML run at the left side of the figure.  In order to diminish the effect of drift detector errors on the system performance, regular checkups are scheduled. If the pipeline is not changed over $Max_{train}$ iterations, the system gives an automatic trigger to start another OAML search.

The OAML framework allows different adaption strategies to update the old model after concept drift has occured. Next, we describe three methods: Basic, Ensemble, and Model Store.

\subsubsection{OAML - Basic}\label{OAML-basic}
The adaptation strategy in OAML - Basic is global replacement, where the old model is completely discarded and a new one is built from scratch. OAML search replaces the previous pipeline with the new one $P^*_{t}$ as soon as possible, and the online learning cycle begins again. In case $(y_t, \hat{y_t})$ does not trigger the drift detector, $(\boldsymbol{x_t}, y_t)$ is used to incrementally train $P^*_{last}$ and the updated pipeline predicts the new sample. OAML - Basic is expected to suit adaptation to high and abrupt concept drift. It is also very scalable with a low memory footprint since old models are discarded and only a limited amount of data is kept in memory at any time. 

\begin{algorithm}
\caption{OAML - Ensemble}\label{algo1}
\textbf{Inputs:} \text{$n_0, n_s, M_a, M_o, t_{max}, S(), D(), (\boldsymbol{X},\boldsymbol{y}), Max_{train}, k $ where $n_0 \geq n_s$} \\
\textbf{Initialization:}
 \text{$OAML_{Search} \Leftarrow S(), E_0 = \small\{\small\}, A_{DD} \Leftarrow$ D(), $t_{train} \Leftarrow 0$}
\begin{algorithmic}[1]
\State $P^*_0 \Leftarrow \mathbf{argmin}_{t_{max}, M_a} OAML_{Search}((x_0,y_0), ..., (x_{n_0}, y_{n_0}))$
\State \text{append $P^*_0$ to $E_0$}
\State $i \Leftarrow n_0 + 1$
\State $t \Leftarrow 0$
\State $A_O \Leftarrow P^*_0$
\While{$x_i \in \boldsymbol{X}$} 
    \Comment{Test-then-train evaluation}
    \State \text{predict $\hat{y_i} \Leftarrow A_O (X_i)$}
    \State \text{evaluate $M_o \Leftarrow M_o(y_i, \hat{y_i})$}
    \State \text{train online model $A_O (X_i, y_i)$ }
    \State \text{update $A_{DD} \Leftarrow M_o$}
    \Comment{Drift detection}
    \If{\text{drift signal $\Leftarrow A_{DD}$ $\lor$ $i - t_{train} \geq Max_{train}$ } }
        \State $t_{train} \Leftarrow t$
        \Comment{Last training}
        \State $t \Leftarrow t+1$
        \Comment{Redesign pipelines}
        \State $X_{sliding} = (x_{i-n_s}, ..., x_i), y_{sliding} = (y_{i-n_s}, ..., y_i) $
        \State $E_t = E_{t-1}$
        \State $P^*_t \Leftarrow \mathbf{argmin}_{t_{max}, M_a} OAML_{Search}(X_{sliding}, y_{sliding}) $
        \State \text{$M_o^E \Leftarrow$ evaluate $E_t (X_{sliding}, y_{sliding})$}
        \Comment{Ensemble evaluation}
        \State \text{$M_o^{P^*} \Leftarrow$ evaluate $P^*_t (X_{sliding}, y_{sliding})$}
        \If{$M_o^E \leq M_o^{P^*}$}
            \If{$A_O = E_t$}
            \State \text{Do nothing}
            \Else
                \State \text{$A_O \Leftarrow E_t$}
            \EndIf
        \Else
            \State \text{$A_O \Leftarrow P^*_t$}
        \EndIf
        \State \text{append $P^*_t$ to $E_t$}
        \Comment{Ensemble update} 
        \If{$ \vert E_t \vert \geq k $}
            \State \text{$E_t$.pop$_{first}$}
        \EndIf
    \EndIf
\EndWhile
\end{algorithmic}
\end{algorithm}

\subsubsection{OAML - Ensemble}\label{OAML-ensemble}
OAML - Ensemble follows the same steps as the basic version except for the adaptation phase. An ensemble of the best performing $k$ pipelines is used to create a dynamic backup ensemble $E = \big\{P^*_i\big\}$. Each pipeline in the ensemble has the same weight and the predictions are aggregated equally. When a drift occurs at time $t$ and OAML is triggered to start a new pipeline search, the output pipeline $P^*_{t}$ is not directly used to replace the old one but instead compared with the backup ensemble, $E_t$, based on their predictive performance over the last sliding window $((\boldsymbol{x_{t-n_s}},y_{t-n_s}), ..., (\boldsymbol{x_t}, y_t))$. If the backup ensemble has a better predictive score than the new pipeline, the current model $A_{O}$ is replaced with $E_t$. The current model $A_{O}$ can be either the ensemble or the previous single pipeline. The reason for this design choice is to allow the model to adjust memory from long term to short term depending on the drift.  The ensemble is updated with the newly found pipeline $P^*_{t}$ and the oldest pipeline in the ensemble is removed in case its length exceeds the limit $k$. This way, the OAML system keeps a memory of previously learned pipelines and the global model replacement strategy is relaxed. Older models can still partially affect the future decisions with their votes in the ensemble until they are replaced. Algorithm - \ref{algo1} shows the steps of OAML - Ensemble in pseudocode, in which performance metrics are supposed to be minimized. 

\subsubsection{OAML - Model Store}\label{OAML-basicstore}
In order to understand the trade-offs of ensembling versus storing models based on their online performance, OAML- Model Store is designed to keep $k$ individual pipelines in memory. This model store ($MS_t$) is effectively a history of the best pipelines used earlier in the data stream. Every time a new pipeline $P^*_t$ arrives, each pipeline in the model store, $P^*_j \in MS_t$, is evaluated with the last $n_s$ samples of data $((\boldsymbol{x_{t-n_s}},y_{t-n_s}), ..., (\boldsymbol{x_t}, y_t))$ and their performances are compared with the new pipeline $P^*_t$. The online model, $A_O$, is updated with the best performing one among the model store pipelines and the new pipeline. If the length of the model store is greater than $k$, the worst performing one is removed and the newest pipeline is added to the store. OAML - Model Store keeps an extended memory similar to OAML - Ensemble, yet the update of this memory is based on individual pipeline performance. Hence, data streams with repeating concepts could benefit from this strategy.


\section{Experiment design}\label{experimentdesign}
In this section, we evaluate our online AutoML system with concept drift data streams, and analyze the results from different perspectives in order to gain a better understanding of the performance of OAML mechanisms. Our code, as well as the data streams and the results of these experiments are publicly available for reproducibility in our github repository.\footnote{\url{https://github.com/openml-labs/gama/tree/oaml}}

\subsection{Data streams}\label{sub:data}
We evaluated our method on 6 well-known data streams from the concept drift literature that are commonly used to test online algorithms' adaptability. Three of these streams are from real-world settings. Others are generated with the online machine learning library MOA \citep{Bifet2011}. Artificial data is critical in online learning research since the existence and characteristics of concept drift can only be certainly known in this setting. Drift is induced in these streams by altering the parameters of the generating function at certain points. Table~\ref{tab:data} presents an overview of all datasets used.

\begin{table}[b]
\caption[data streams]{Overview of the data streams used in our experiments\footnotemark[1]}
\label{tab:data}
\resizebox{\textwidth}{!}{%
\begin{tabular}{p{2.2cm}p{1cm}p{1cm}p{9cm}}
\toprule
\textbf{Data stream} & \textbf{Samples} & \textbf{Features} & \textbf{Data generation} \\ \toprule

Electricity \citep{Harries1999} & 45312 & 7 & A time series electricity price data from the Australian New South Wales market. The prices are volatile and change with demand and supply every five minutes. The class shows the price direction per day average.\\
\hline
Airlines \citep{Bifet2011} & 539383 & 9 & A time series data with each sample representing a flight with its schedule information.The class shows whether the flight arrived on time. Flight schedule changes daily or weekly.\\
\hline
Vehicle \citep{Duarte2004} & 98528 & 100 & Sensor data from a wireless sensor network for moving vehicles. The class divides the vehicles into categories. \\
\hline
SEA Abrupt \citep{Street2001} & 500000 & 3 & Data generated with Streaming Ensemble Algorithm (SEA). Abrupt drift is introduced at the middle point by shifting from one classification function to the other. The drift window is set to 1 and 10\% label noise is added. Drift magnitude is set to high by switching between most different functions.\\
\hline
HYPERPLANE Gradual \citep{Hulten2001} & 500000 & 10 & Data generated with Rotating Hyperplane Algorithm. Drift is added by shifting the orientation and position of the hyperplane. This shift creates a gradual drift by setting the window size to 100000. Data includes 5\% label noise.\\
\hline
SEA Mixed \citep{Street2001} & 500000 & 3 & Data generated with Streaming Ensemble Algorithm (SEA). Abrupt drift is introduced the same as SEA-Abrupt data. In addition, gradual drift is added before and after abrupt drift with window of 100000 samples. Data includes 5\% label noise. \\ \bottomrule
\end{tabular}%
}
\footnotetext[1]{Source: Search \url{www.openml.org} for datasets tagged `concept\_drift`.}
\end{table}

\subsection{OAML configuration}\label{sub:setup}
OAML can be run with different settings that can be adjusted based on the application requirements or preferences. The online learning system we assume relaxes the single pass assumption since we do keep a sliding window of data for pipeline search. Still we do restrict kept memory against the infinite flow of data.  The initial batch size, $n_0$, and sliding window size, $n_s$, determine how much data is fed into the automated pipeline design process. In this paper, we use $5000$ for both values as a result of initial ablation studies with different streams. Optimal data memory is related to the existence of drift in data, the storage capacity and speed requirements of the system, and its effect on the search algorithm performance. Hence, they should be considered for application specific selection. The AutoML budget, $t_{max}$, is set to $600$ seconds, as aligned with the batch size. We prioritize predictive performance over speed yet OAML can further be adjusted for a faster system by parallel processing of the online learning and automl modules. The online learning module can be kept running with the anytime model while automl searches for a better pipeline. OAML allows choosing among several online learning metrics. In our experiments, the performance metric is set to prequential accuracy for both the pipeline search ($M_a$) and online learning phase ($M_o$). As for the search algorithm, we mainly use an evolutionary search algorithm for evaluating the performance of OAML because of its adaptation capability with drifting data \citep{Celik2021}, yet we also conduct experiments to compare different choices for the search algorithm. The drift detection algorithm is set to EDDM due to its precision in detecting both gradual and abrupt drift, as explained in Section~\ref{OnlineLearning}. In order to decrease our dependency on the drift detector, the alternative trigger for pipeline search, $Max_{train}$ is set to $50000$ samples, considering the size of the data streams. 

\subsection{Baselines and state of the art}\label{sub:baselines}
We compare our method against the most competitive alternative techniques in this area. First, we compare against Leveraging Bagging \citep{Bifet2010}, which has shown to outperform many other online learning techniques in the literature. We also include the Hoeffding Adaptive Tree \citep{Hulten2001} as one of the strongest non-ensembling techniques. Finally, we also compare against the state-of-the-art AutoML library for online learning, ChaCha \citep{Wu2021}. Even though it does not construct entire pipelines, and includes a much smaller model search space than OAML, it should be very competitive on the datasets in this study since these datasets don't require significant preprocessing.

\section{Results}\label{sub:results}
Here we present the results of our experiments. We first compare the different versions of our method with each other and with the baselines described above, on both real-world and synthetic datasets. Next, we analyse which pipelines are actually generated by OAML, which components they contain, and whether the actively used pipelines come directly from the AutoML phase, or from the ensemble or model store. Finally, we also explore which AutoML search algorithms perform best for the different types of concept drift.

\subsection{OAML experiments with artificial data}\label{artificialdata}

\begin{figure*}[]
\centering
\subfigure[][]{\includegraphics[trim=200 30 120 0,width=1.90in]{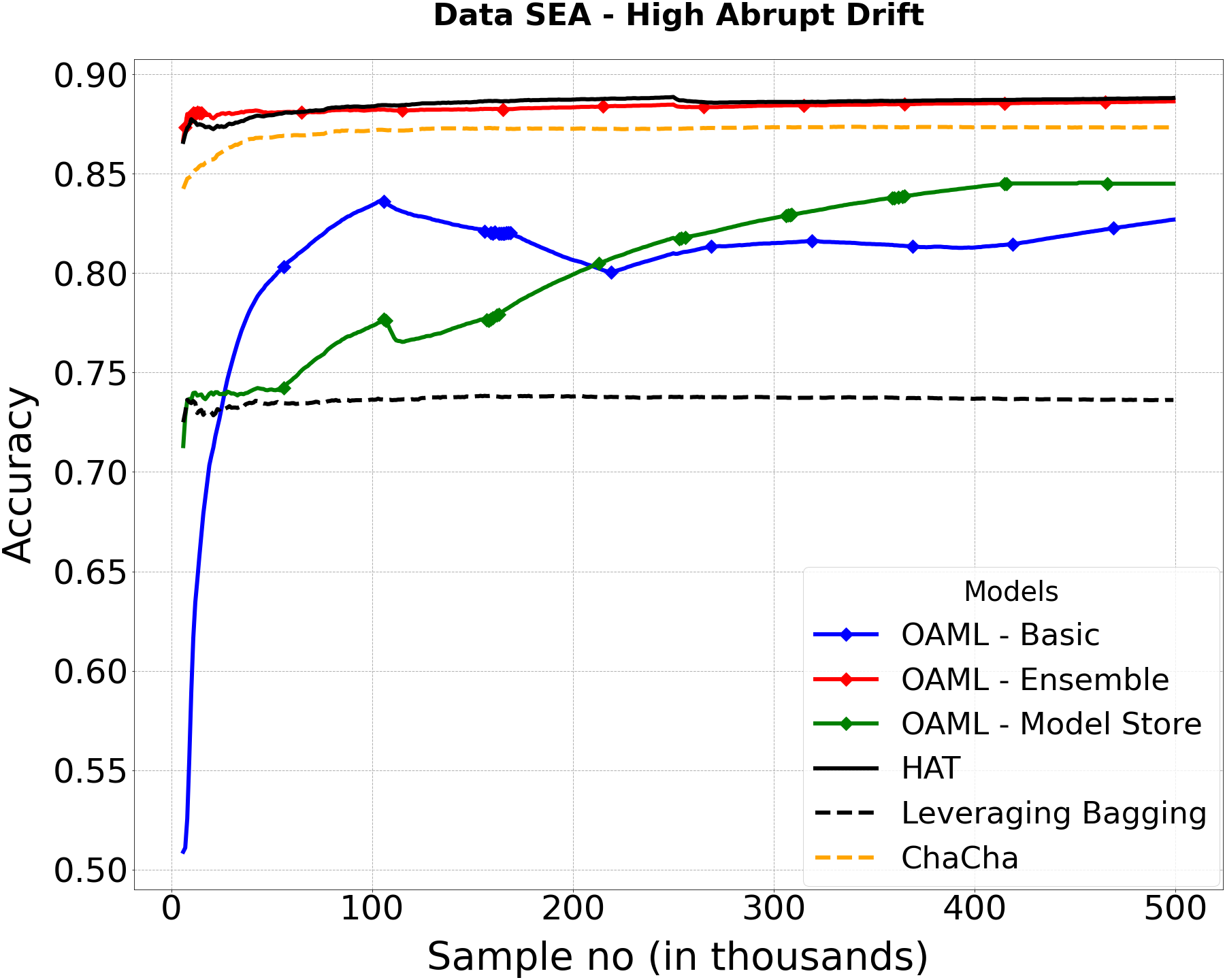}%
}
\hspace{1.5cm}
\vspace{0.3cm}
\subfigure[][]{\includegraphics[trim=200 30 120 0,width=1.90in]{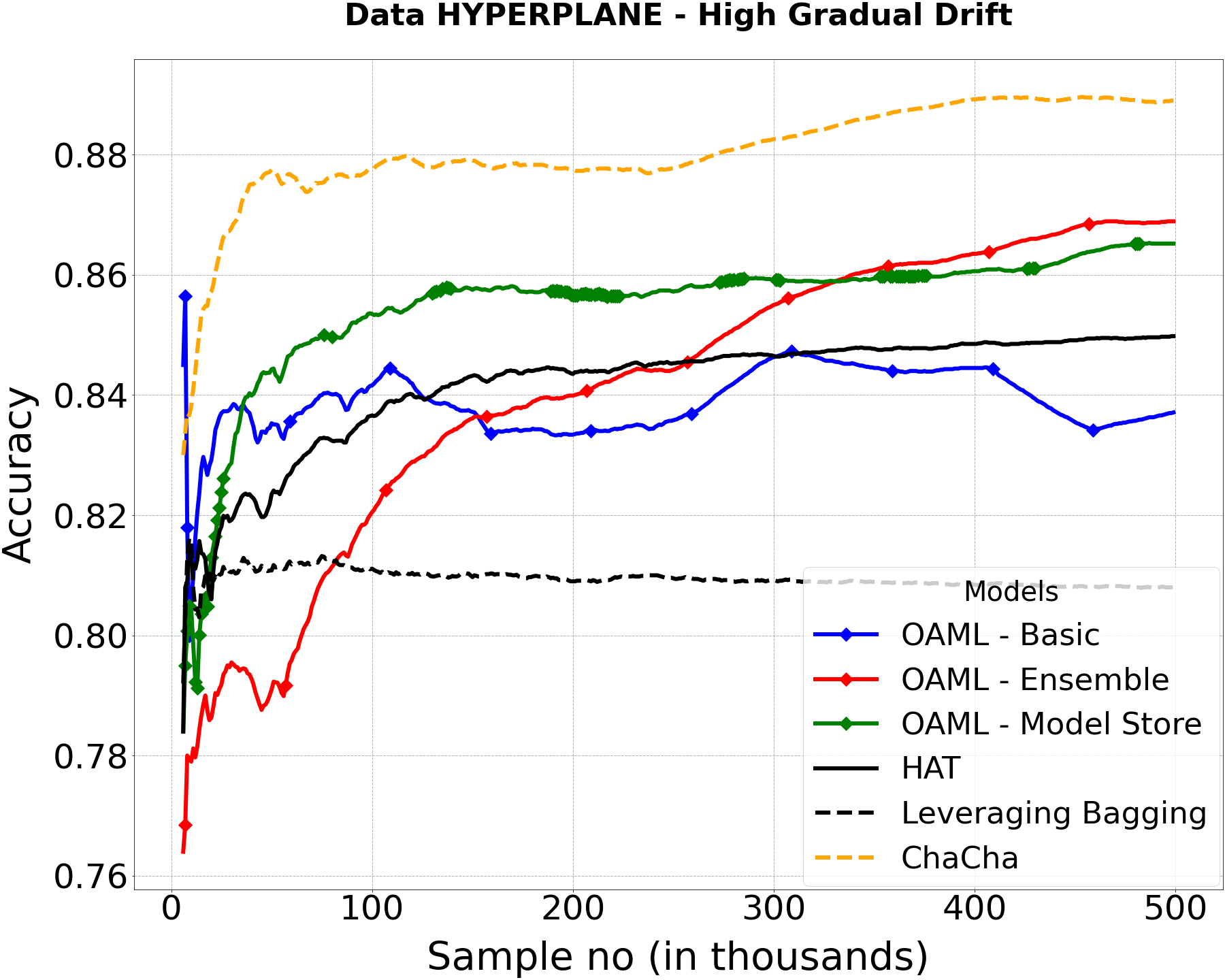}%
}
\hspace{1.5cm}
\vspace{0.3cm}
\subfigure[][]{\includegraphics[trim=200 30 120 0,width=1.90in]{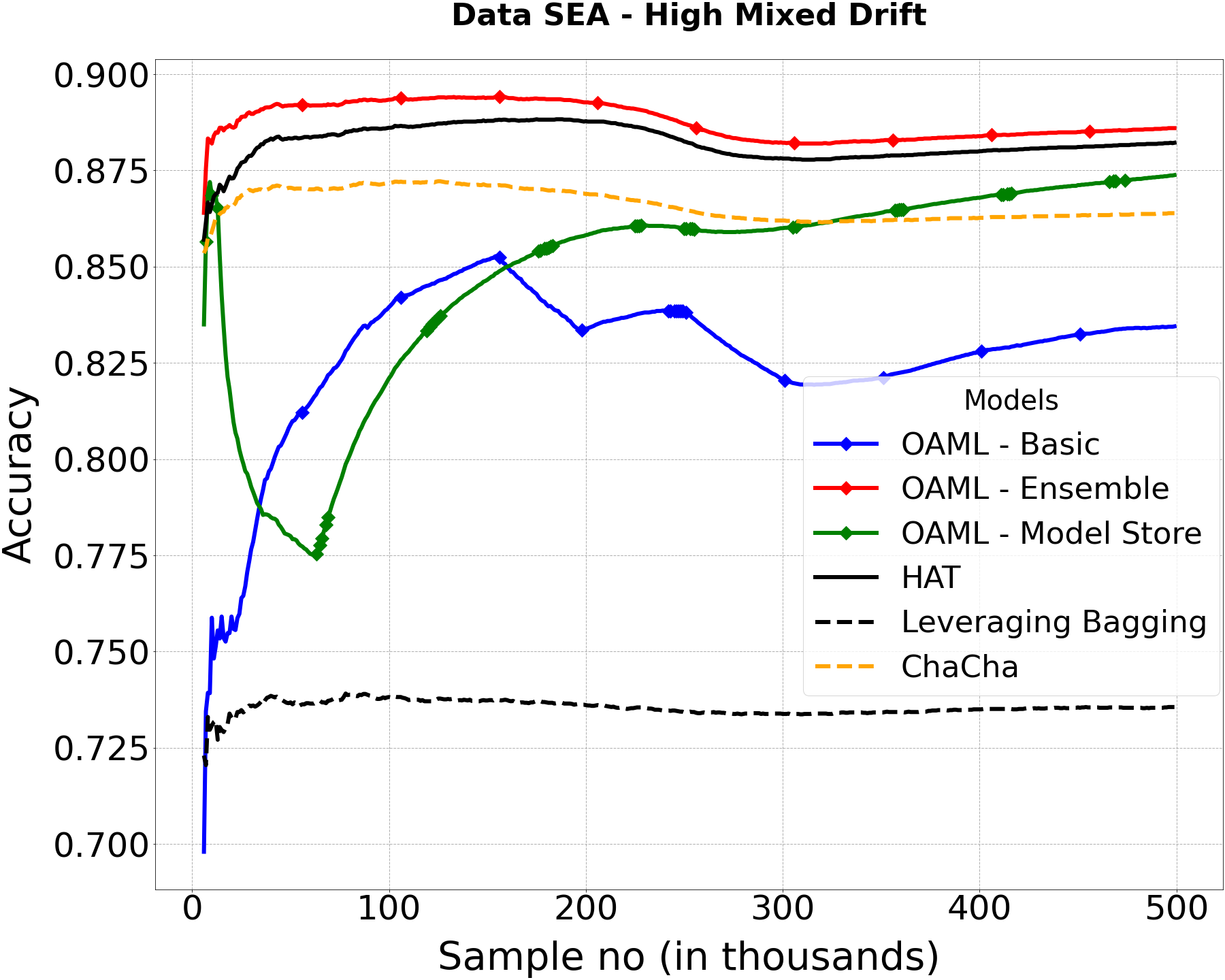}%
}
\caption{Prequential performance for artificial data streams:
(a) SEA - High abrupt drift
(b) HYPERPLANE - High gradual drift
(c) SEA - High mixed drift
}
\label{fig:artificial}
\end{figure*}

Next, we evaluate the adaptability of the OAML framework on data streams with artificially controlled drifts, leading to both abrupt and gradual shifts in the underlying concepts. Figures~\ref{fig:artificial}-a to \ref{fig:artificial}-c plot the results of data streams with gradual, abrupt and mixed drift, respectively. As explained in Table~\ref{tab:data}, each drift is created with a high magnitude to see their effects more clearly. Drift and retraining points are again shown with markers. The plots of each OAML version include markers at the known drift points of the streams, showing that models are retrained after the introduced drift points as expected. For SEA - High Abrupt Drift data stream (\ref{fig:artificial}-a), OAML - Ensemble and HAT exhibit similar accuracy levels, followed closely by ChaCha, with almost no drop in performance at the middle drift point. This shows the fast adaptation capability of OAML since the SEA data has a sudden and significant concept change. OAML - Basic, Model Store and Leveraging Bagging cannot reach that level of adaptability. 

For Hyperplane - High Gradual Drift data, ChaCha performs best, suggesting that its model search technique (quite different from the evolutionary approach used here by OAML) seems to perform well under gradual drift. This suggests that it would be interesting to integrate it in OAML as well. OAML - Model Store outperforms most other methods, showing a significant capability to adapt fast when drift is introduced slowly. OAML - Model Store uses predictions of a single pipeline in a set of pipelines fit to prior data. This is particularly fast adapting when the drift detector is triggered multiple times that updates the set quickly as it is the case at the beginning of the data. Unlike Model Store, OAML - Ensemble requires more than a single best pipeline and likely struggles to form a good combination of pipelines with the data distribution shifting continuously. Since also OAML-Ensemble catches up but OAML-Basic does not, keeping a certain amount of memory seems to be work better under gradual drift.

Figure~\ref{fig:artificial}-c shows the results of SEA - High Mixed Drift, which is designed to combine the challenges of both drift types and identify the approach that handles that the best. In this case, although quite close to HAT, OAML - Ensemble performs best across the whole data stream.  OAML - Ensemble has the advantage of an average prediction of a diverse set of learners trained with different segments of prior data. Since the initial part of data evolves with a gradual drift, these base learners are fit to various distributions. Likely, the diversity helps the overall model to create a robust prediction least affected by abrupt shifts. It can also be seen that OAML - Model Store first suffers, but recovers quite well in the second half of the stream with an increase in performance with every concept drift detected. 

Overall, we see that, again, OAML - Ensemble can adapt to different types of drift and outperform baselines while the speed of adaptation differs with the drift type. It can particularly handle sudden drift points better than the other versions, likely by combining the average predictions of a diverse set of pipelines, which themselves may include ensemble learners. 

\subsection{OAML experiments with real-world data}\label{realdata}

\begin{figure}
    \centering
    \vspace{1cm}
    \includegraphics[scale=0.1]{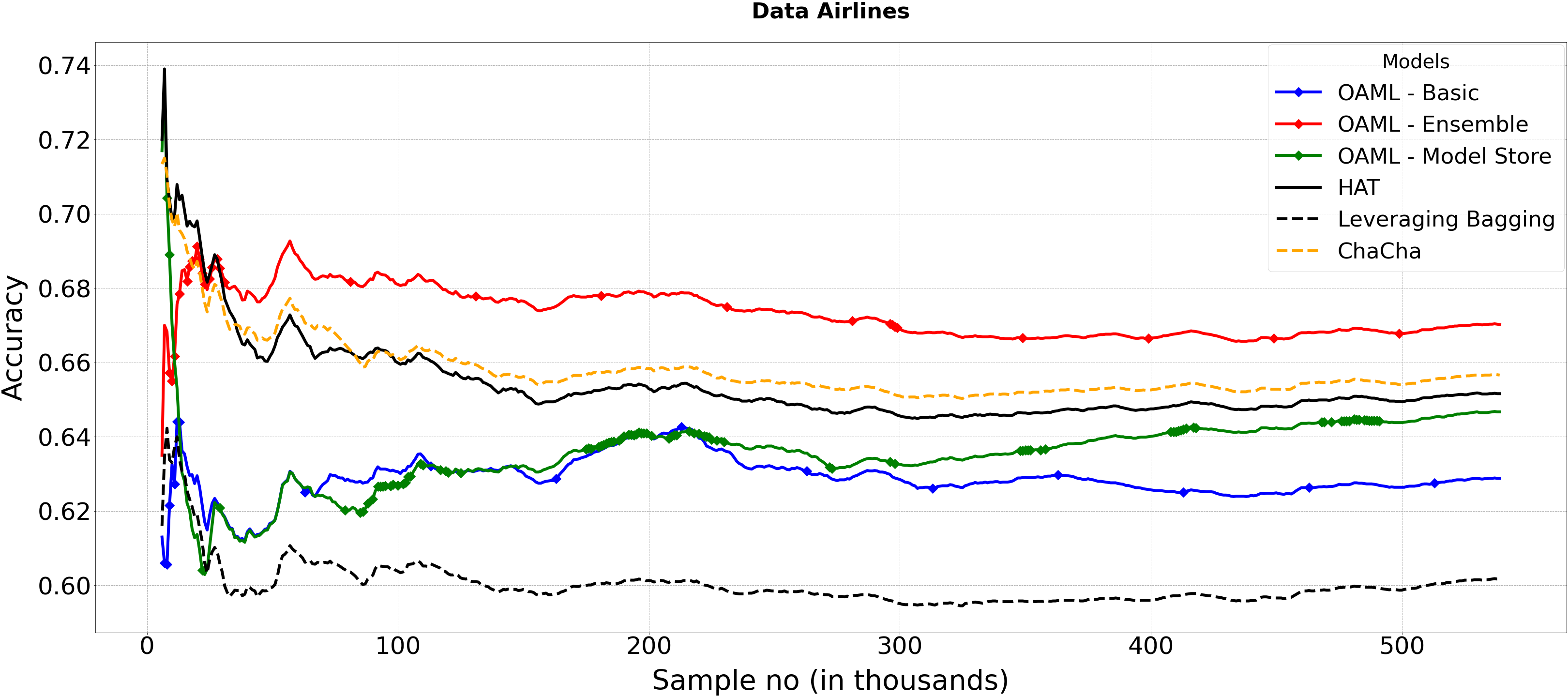}
    \caption{Prequential performance for Airlines data stream}
    \label{fig:airlines}
    \centering
    \vspace{1cm}
    \includegraphics[scale=0.1]{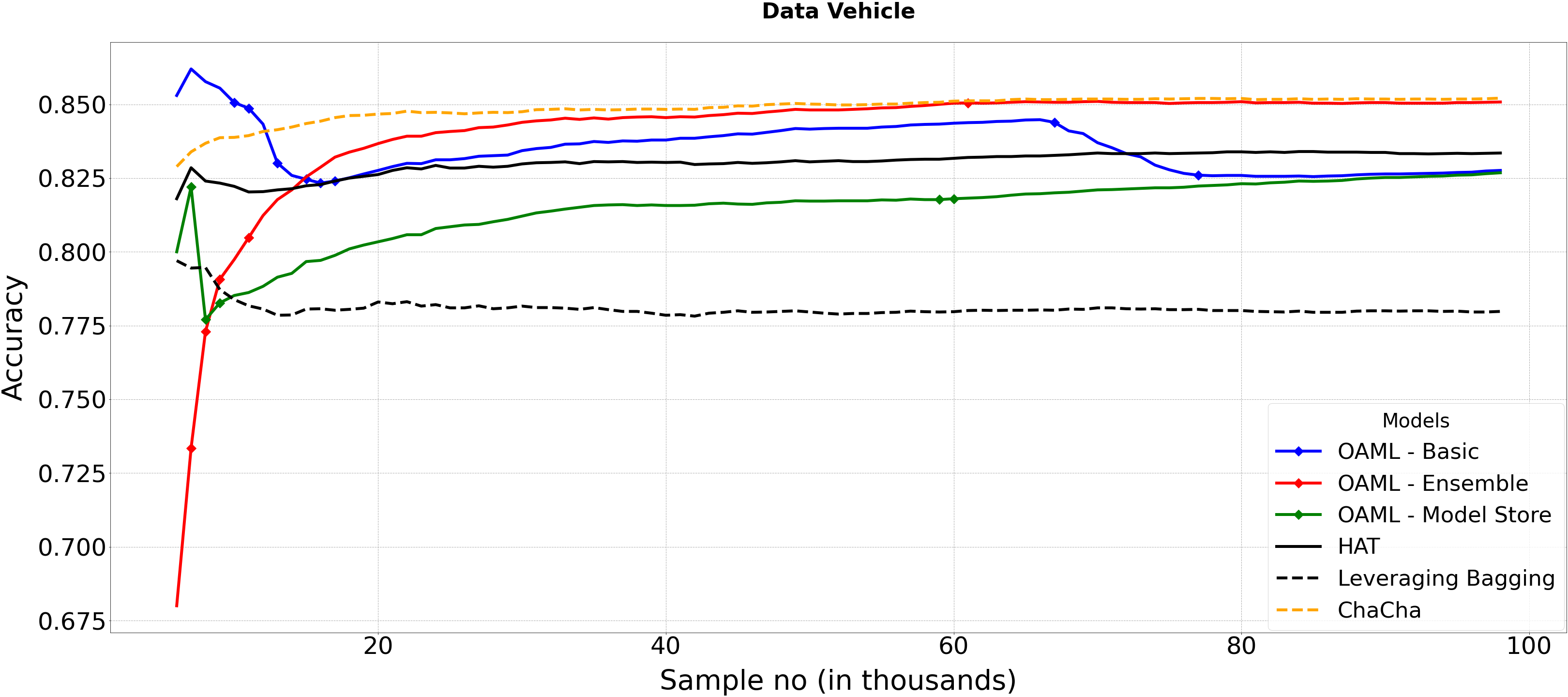}
    \caption{Prequential performance for Vehicle data stream}
    \label{fig:vehicle}
    \centering
    \includegraphics[scale=0.1]{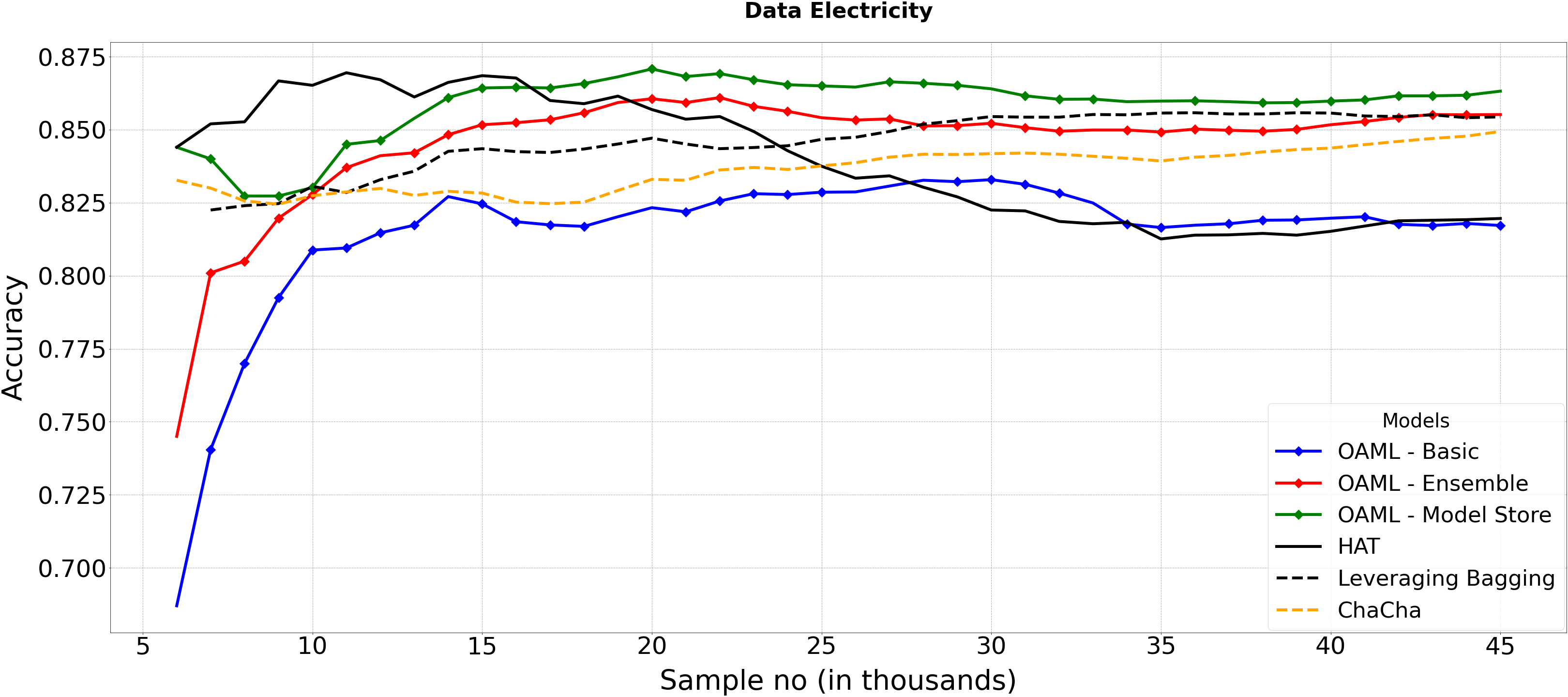}
    \caption{Prequential performance for Electricity data stream}
    \label{fig:electricity}
\end{figure}

Evaluating OAML's capability to handle real-world challenges is critical to assess its practical utility. 
The results on the real data streams are shown in Figures~ \ref{fig:airlines}, \ref{fig:vehicle} and \ref{fig:electricity}. Each line plots the prequential accuracy of a different algorithm over the entire stream. The markers on the plot lines indicate drift and retraining points, i.e. that either the drift detector triggered an alarm at that point in the stream or it is regular retraining point.  

For the Electricity data stream (Figure~\ref{fig:electricity}), drift is detected often due to the quickly changing nature of data from day to day. After a couple of initial iterations, OAML with either an ensemble or model store outperforms the remaining methods. The Model Store strategy may benefit from cyclic effects in this data stream. OAML - Ensemble performs best throughout the airlines (Figure~\ref{fig:airlines}) and vehicle (Figure~\ref{fig:vehicle}) data streams, although it is tied with ChaCha on the latter, which seems to have very little or very gradual concept drift.

Baseline online learners perform relatively good in Electricity data stream though they fail to reach the level of OAML - Model Store.  Note that we ran baselines with their default configuration on all data streams. It is likely that optimizing them to each stream individually would yield better results. In fact, as shown in Section \ref{pipeline}, OAML does exactly this: it often uses a (tuned) Leveraging Bagging pipeline, while switching to HAT in other parts of the stream. This underlines that, not surprisingly, the AutoML tuning typically yields a significant improvement over untuned algorithms, and that OAML manages to bring these benefits to dynamic environments.

Overall, OAML handles concept drift complexities that are sourced from different data generating environments quite well, especially with the backup ensemble strategy.

\subsection{Pipeline analysis}\label{pipeline}
In this section, we analyze the pipelines designed and updated while running OAML.

\begin{figure}
    \centering
    \includegraphics[scale=0.14]{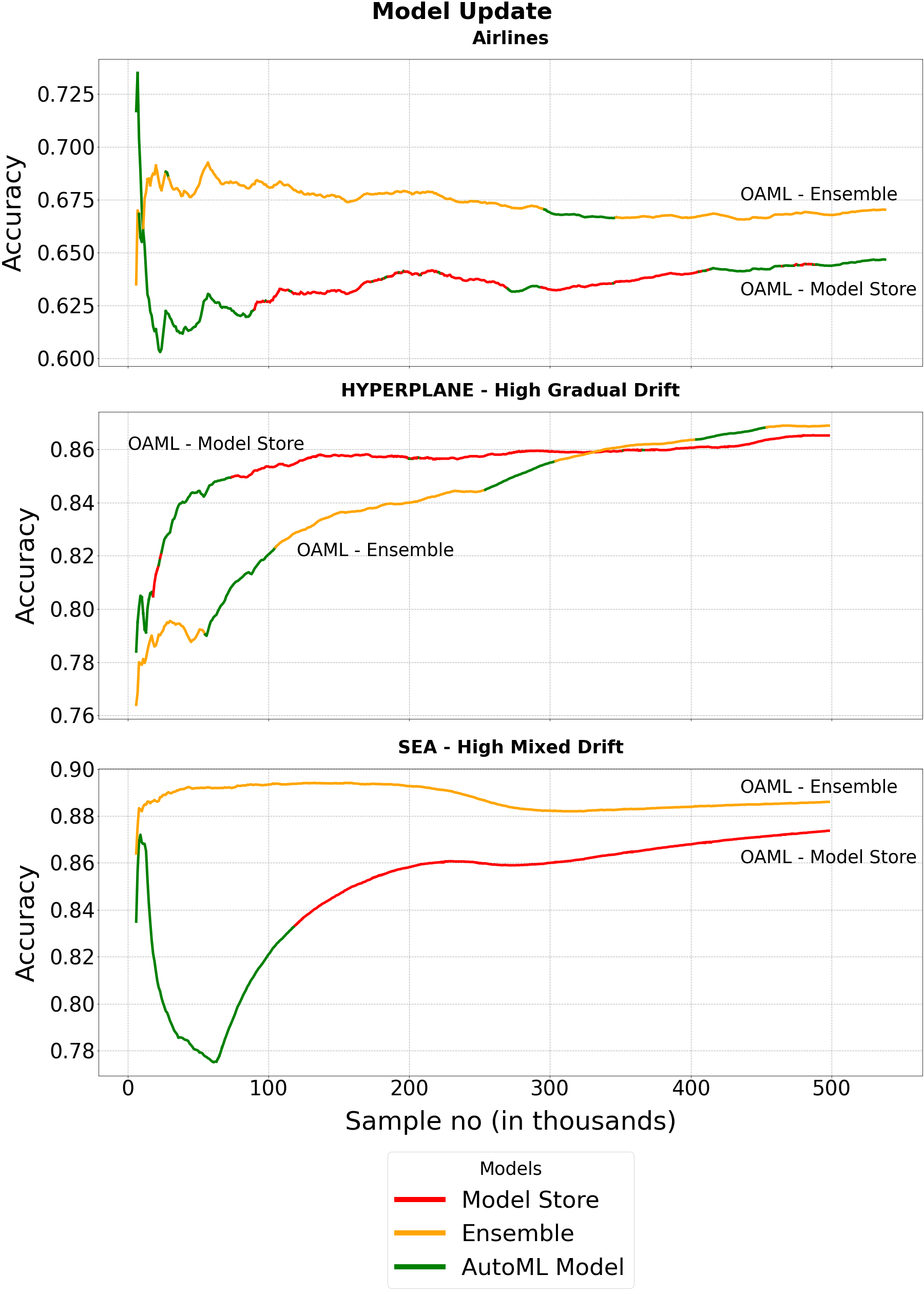}
    \caption{Model update switches for \textit{OAML - Ensemble} and \textit{Model Store}}
    \label{fig:modelupdate}
\end{figure}

First, we analyze whether the actively used pipeline comes directly from the AutoML optimizer, or whether it is recovered from the Model Store or the backup Ensemble. OAML can decide to switch between them according to what seems best. Figure~\ref{fig:modelupdate} shows the switch points between these models throughout the data streams Airlines, Hyperplane-Gradual and SEA-Mixed. Since OAML starts with a pipeline created by the \textit{AutoML Model}, all lines start as green. OAML-Ensemble tends to switch quickly to the ensemble model (yellow), except for Hyperplane-Gradual, where the ensemble is occasionally replaced by a new AutoML model. The same holds for OAML-Model Store, although here it is the Airlines data stream where new AutoML models are frequently injected. For SEA-Mixed Drift, we see that the sudden recovery of OAML-Model Store, previously seen in Figure~\ref{fig:artificial}-c, is due to a switch from the old AutoML model to the Model Store (red).

Gradual drift leads to more switches for both versions than the Mixed drift setting where drift occurs with varying speeds. In that case, OAML keeps using the \textit{Ensemble} or \textit{Model Store} options which handle these variations better. This also shows the benefit of pipeline redesign since the models only switch to \textit{AutoML} when the newly redesigned pipeline is better than the Ensemble or all the pipelines in the Model Store.

\begin{figure}
    \centering
    \includegraphics[scale=0.14]{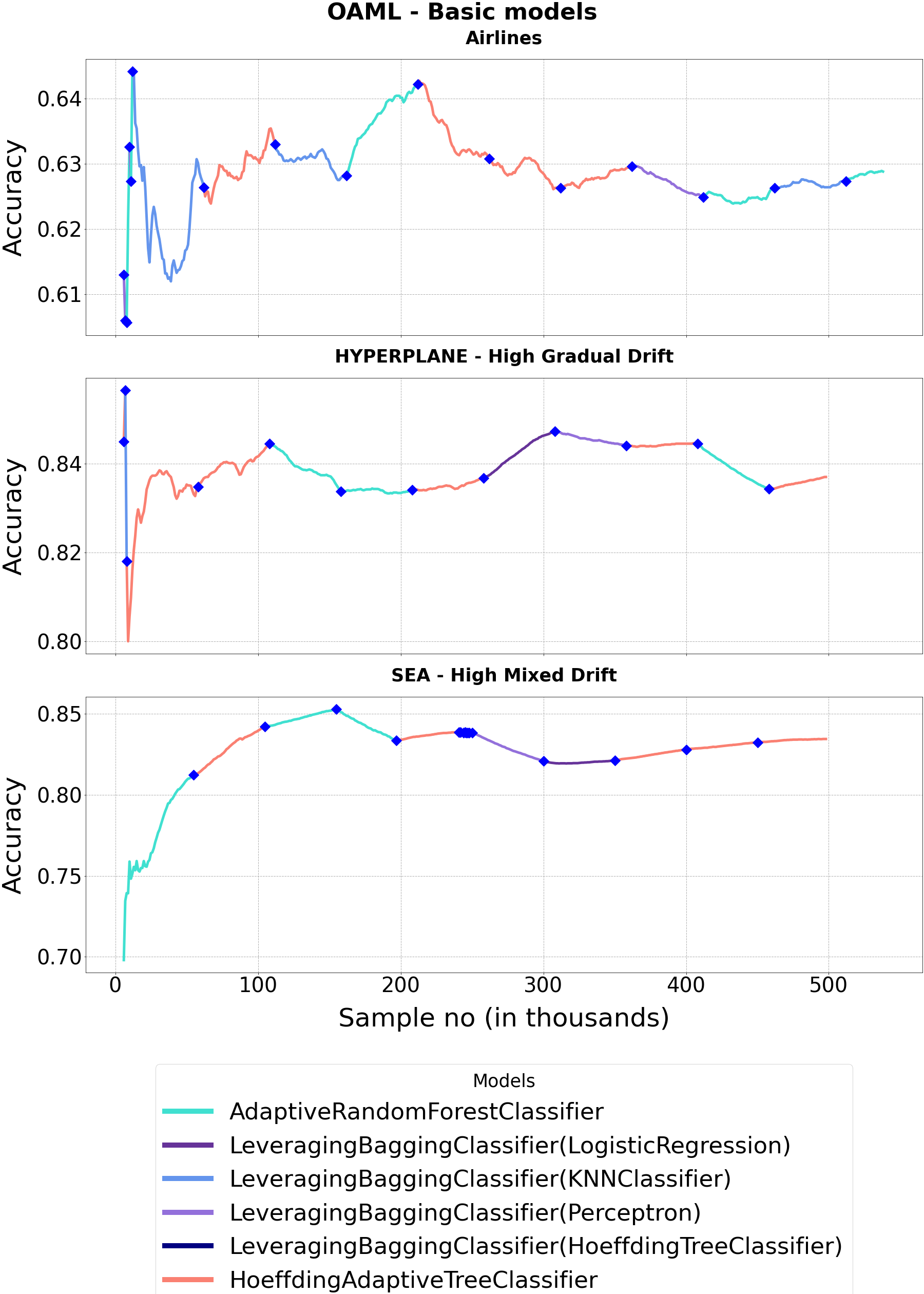}
    \caption{Models used by \textit{OAML - Basic} throughout streams}
    \label{fig:models}
\end{figure}

Looking deeper into the pipeline optimization phase, we examine exactly which models are used throughout the online learning process. Figure~\ref{fig:models} shows how the used classifier switches throughout the stream when using OAML-Basic, with the retraining points marked in between. 

It can be seen that each data stream experiences several model switches, and that the model used can be an online ensemble or single online learner. Although ensemble learners are more dominant, the Hoeffding Adaptive Tree (HAT) is also used in each data stream at several points. For Hyperplane-Gradual data, changes between models are relatively slow, which reflects the gradual drift effect on algorithm selection. With the SEA-Mixed stream, we also see quicker jumps at the abrupt drift points in the middle of the stream. This is the phase where OAML tries to find a fitting model to the new concept after a quick change. When Figure~\ref{fig:models} shows the same color line segments before and after a retraining point, this means that its hyperparameters were retuned instead of being replaced with a new model. This can be observed in all three data streams with the HAT or ARF classifiers. 

\begin{figure}
    \centering
    \includegraphics[scale=0.14]{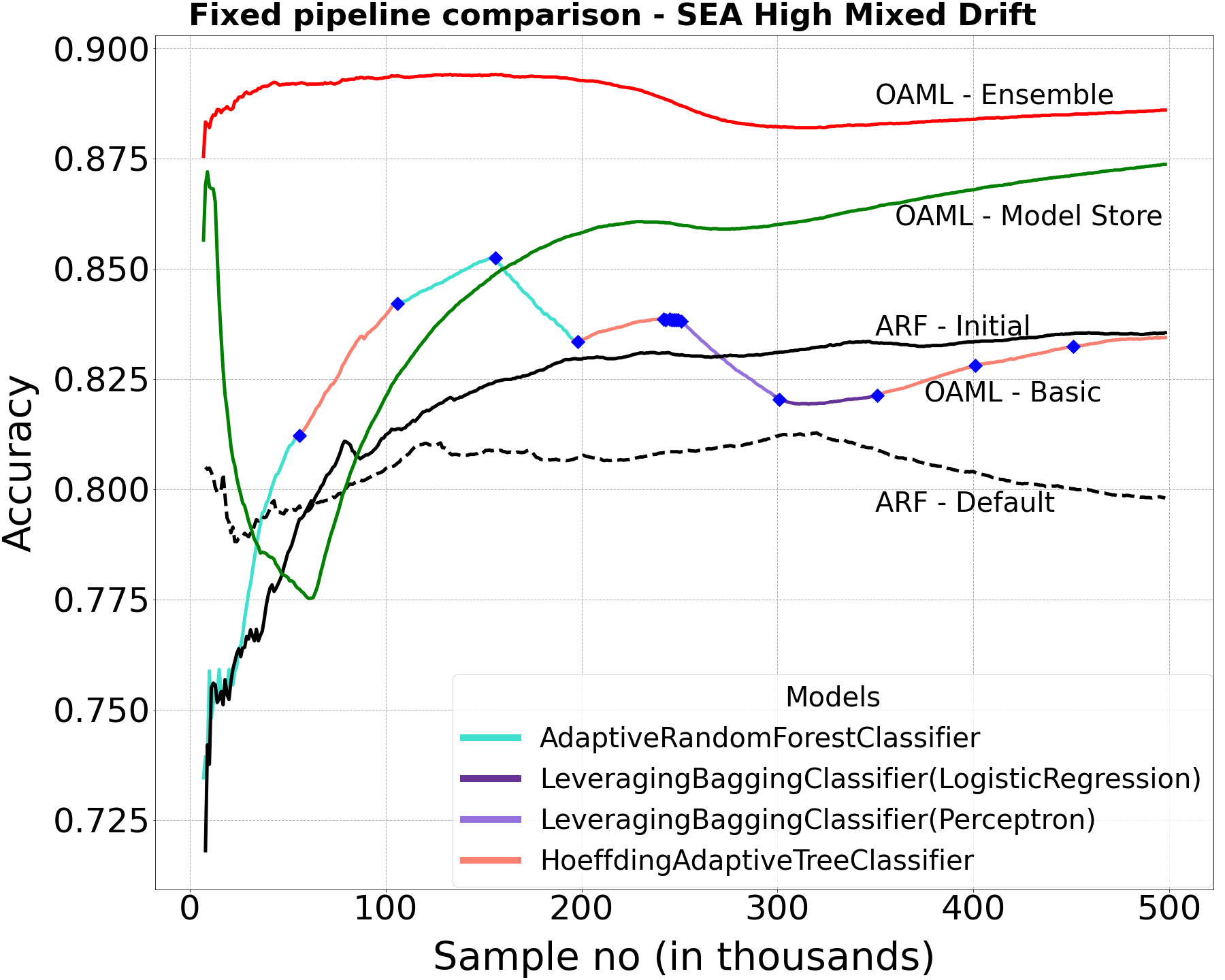}
    \caption{Comparison of fixed initial OAML pipeline with \textit{OAML full runs}}
    \label{fig:FIP}
\end{figure}

Furthermore, we evaluate what happens if we keep the initial pipeline found by OAML active throughout the data stream instead of re-optimizing pipelines at the drift points. Figure~\ref{fig:FIP} shows the results of this analysis for SEA-Mixed Drift data, where the initial pipeline used as Adaptive Random Forest (ARF - Initial shown by solid black line) by OAML - Basic. As a baseline, ARF with default hyperparameters is also included (ARF - Default shown with black dashed line). Although ARF fluctuates in performance for different applications due to the randomness inhibited in the model, overall it can be seen that the fixed initial pipeline fails to reach the level of adaptive OAML in performance as data starts to drift. Yet, it is quite robust and doesn't get affected by the sudden drift point at the middle as much as the OAML - Basic's chosen Leveraging Bagging classifier. Looking at the overall performance of the OAML versions, it is still clear that re-optimizing pipeline design is likely to outperform the initially optimized one through a drifting data stream. Fixing the initial pipeline (\textit{Train once} strategy) is also found to be dominated by re-optimizing the pipelines strategy in \cite{Celik2021}. 

Overall, our  analysis shows that both hyperparameter tuning and algorithm selection are used interchangeably by OAML, indicating that pipeline redesign can lead to a better performing model as shown in the analysis of fixed pipelines. 

\subsection{AutoML Search Algorithm Effect}\label{searchalgorithm}
\begin{figure}
    \includegraphics[scale=0.1]{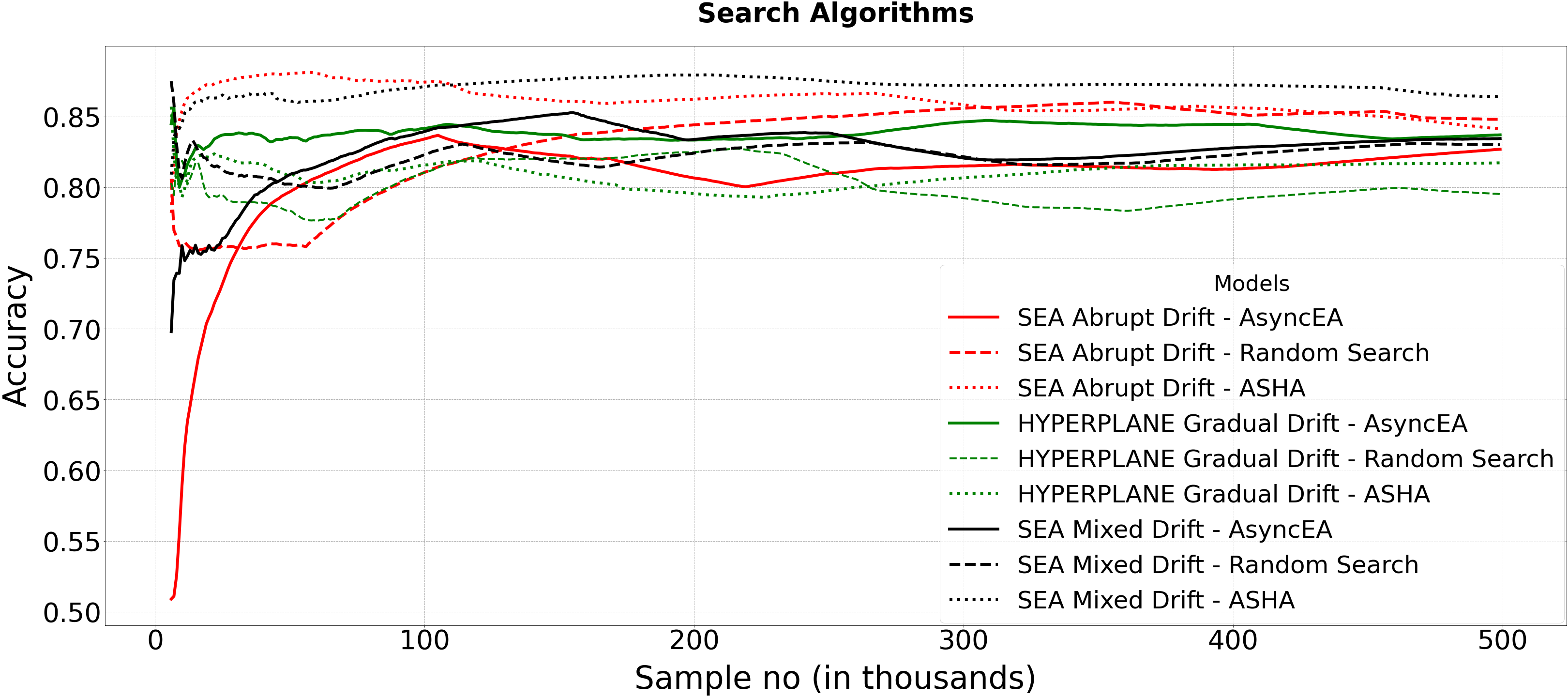}
    \caption{Performance comparison of search algorithms \textit{Random Search, ASHA and AsyncEA}}
    \label{fig:searchalgs}
\end{figure}

In order to understand how the choice of search algorithm impacts drift adaptation in OAML, we experimented with random search, asynchronous successive halving (ASHA) and evolutionary algorithm (AsyncEA) on artificial data streams with different drift types. This selection is based on prior research results (\cite{Celik2021}) and can be extended with different search algorithms. The results are shown in Figure~\ref{fig:searchalgs}, where each color represents a data stream and each line type a different search algorithm. 

For SEA generated data, ASHA performs slightly better compared to others (red dotted line), especially in the beginning of the stream. This could be due to the successive halving strategy working well with quick adaptation to abrupt drifts present in these data streams. ASHA will evaluate more pipelines randomly but quickly, while the evolutionary approach may need more iterations to start evolving good pipelines. On the other hand, ASHA performed less well on Hyperplane-Gradual, where more continuity is needed in the pipelines to follow the gradual change, and hence evolutionary search algorithm adapts better (green solid line plot). Overall, it can be observed that each algorithm adapts quite well without a drastic drop in performance. This is aligned with previous findings \citep{Celik2021} that although some search algorithms fit better to specific drift types with differences in computational cost, search algorithm selection doesn't greatly affect the adaptation capability as much as the other algorithm design choices in online learning pipeline search. 

\section{Conclusion}\label{Concl}
We introduced a novel framework that enables AutoML methods to be applied effectively in dynamic environments with evolving data streams. It automatically searches for optimal pipelines that can contain preprocessing techniques and that exclusively use online learning algorithms which adapt to gradual changes in the data. It also detects concept drift and can automatically redesign or retune the pipelines when needed. As a result, it addresses both the traditional challenges of AutoML, such as combined algorithm selection and hyperparameter optimization under time constraints, as well as new challenges particular to real-world data streams, such as concept drift, memory restrictions, and forgetting mechanisms. 

We defined a rich pipeline search space that includes many online learning algorithms, ensembles, data and feature preprocessing steps. Our framework includes three strategies to update the currently used pipelines after drift is detected: \textit{Basic}, which fits a new pipeline every time drift is detected and replaces the old one; \textit{Model Store}, which keeps a memory of the best performing prior pipelines and selects the best current one; and \textit{Ensemble}, which makes a backup ensemble from the best prior pipelines.

We evaluate the developed method on both real-world and artificial data streams with concept drift and compare its performance with several baselines and state-of-the-art systems. The results show that OAML-Ensemble performs consistently well on data with various kinds of concept drift, while OAML-Model Store performs best when there are cyclic/seasonal processes underlying the data stream. We also examine how OAML behaves by tracking the underlying changes in the generated pipelines through time. Our results show that there is no online algorithm that is optimal across the life cycle of a data stream. As the data changes over time, online learners in these pipelines are either replaced, or their hyperparameters are re-optimized to adapt to the changing data. This demonstrates the benefits of automating both algorithm selection and hyperparameter tuning for online learning, in contrast to previous studies that focused only on either one of these. Currently, OAML budgets a process time for the pipeline search, which is not needed in the baseline online learners. Yet, it is possible to advance OAML by eliminating the need for the initial waiting time for the first pipeline search and applying a continuous search algorithm that can handle data flow by sample instead of batch. In addition, currently available public data streams do not suffice to understand the importance of preprocessing in the search space, which requires further collection of imperfect, concept drift data.  In all, we believe that this work opens up interesting avenues for further research.


\newpage
\bmhead{Acknowledgments}

We would like to give special thanks to Pieter Gijsbers for his help in integrating OAML into the GAMA library. This research was supported by the Dutch Foundation for Scientific Research (NWO) under the DACCOMPLI grant, and by the European Commission's H2020 program under the StairwAI grant. It was also partially supported by TAILOR, a project funded by EU Horizon 2020 research and innovation programme under GA No 952215.

\section*{Declarations}

\begin{itemize}
\item Funding: The research is funded under the NWO project DACCOMPLI and partially by TAILOR project. 

\item Availability of data and materials and code: All our data, materials and code are available publicly at \url{https://github.com/openml-labs/gama/tree/OAML}. 

\item Financial or non-financial interests: Not applicable.

\item Ethics approval: Not applicable.

\item Consent to participate: Not applicable.

\item Welfare of animals: Not applicable.

\item Authors' contributions: The authors contributed equally to the research and publication.

\item Conflicts of interest/Competing interests: Not applicable.

\item Consent for publication: Not applicable.
\end{itemize}





\bibliography{references}


\end{document}